\documentclass[sigplan,screen]{acmart}
\renewcommand\footnotetextcopyrightpermission[1]{} % removes footnote with conference information in first column
\usepackage{multirow}
\usepackage[misc]{ifsym}
\usepackage{fancyhdr}
\AtBeginDocument{%
  \providecommand\BibTeX{{%
    \normalfont B\kern-0.5em{\scshape i\kern-0.25em b}\kern-0.8em\TeX}}}

\setcopyright{acmcopyright}
\copyrightyear{2018}
\acmYear{2018}
\acmDOI{}

\begin{document}

\title{MBDF-Net: Multi-Branch Deep Fusion Network for 3D Object Detection}

\author{Xun Tan\textsuperscript{1}, Xingyu Chen\textsuperscript{1}, Guowei Zhang\textsuperscript{1}, Jishiyu Ding\textsuperscript{2},  Xuguang Lan\textsuperscript{1, \Letter}}
\thanks{{\Letter} Corresponding author}
\affiliation{
      \institution{Institute of Artificial Intelligence and Robotics, Xi'an Jiaotong University\textsuperscript{1}, the Second Academy of CASIC\textsuperscript{2}}
%   \city{Xi'an}
  \country{China}}
\email{tanxun31@stu.xjtu.edu.cn, xingyuchen1990@gmail.com, zgw1119@stu.xjtu.edu.cn}
\email{dingjishiyu@126.com, xglan@mail.xjtu.edu.cn}

% \author{}
% \affiliation{%
%   \institution{Institute of Artificial Intelligence and Robotics, Xi'an Jiaotong University,xglan@mail.xjtu.edu.cn}
%   \city{Xi'an}
% \email{}

\renewcommand{\shortauthors}{}

\begin{abstract}
  Point clouds and images could provide complementary information when representing 3D objects. Fusing the two kinds of data usually helps to improve the detection results. However, it is challenging to fuse the two data modalities, due to their different characteristics and the interference from the non-interest areas. To solve this problem, we propose a Multi-Branch Deep Fusion Network (MBDF-Net) for 3D object detection. The proposed detector has two stages. In the first stage, our multi-branch feature extraction network utilizes Adaptive Attention Fusion (AAF) modules to produce cross-modal fusion features from single-modal semantic features. In the second stage, we use a region of interest (RoI) -pooled fusion module to generate enhanced local features for refinement. A novel attention-based hybrid sampling strategy is also proposed for selecting key points in the downsampling process. We evaluate our approach on two widely used benchmark datasets including KITTI \cite{Geiger2013VisionMR} and SUN-RGBD \cite{sunrgbd}. The experimental results demonstrate the advantages of our method over state-of-the-art approaches.
\end{abstract}





\maketitle

\section{Introduction}
Object detection is an important topic in the computer vision community which has been widely used in many realistic scenarios. With the development of deep learning technologies, a variety of image-based methods  \cite{Ren2015FasterRT, Liu2016SSDSS, Redmon2016YouOL, Tian2019FCOSFC} have been proposed, which makes great progress in 2D object detection. However, 2D detection methods can seldom provide accurate results in 3D space due to the lack of depth information. That is problematic in many tasks such as automatic driving and intelligent robot. Therefore, 3D object detection \cite{Guo2020DeepLF} attracts more attention in recent years.

Point cloud is naturally suitable for 3D object detection since it directly provides 3D information and has strong robustness. However, point cloud is usually sparse, disordered and also lacks texture information. It is easy to be disturbed by objects with similar shapes, which limits the accuracy of detection. Images can provide color and texture information but lack depth information. Directly using images for 3D object detection is also bothered by many problems such as occlusion and deformation, which makes it extremely vulnerable to the influence of the environment.

Obviously, combining the two kinds of data and making good use of the advantages of each one could greatly help to improve the detection results. Therefore, many camera and LiDAR fusion-based detection methods have been proposed, which can be roughly grouped into two lines, including result-level fusion methods and feature-level fusion methods. The result-level fusion methods \cite{Qi2018FrustumPF, 2019Fconvnet, Pang2020CLOCsCO} handle each modality independently and then integrate the corresponding results. But these methods do not fully utilize the complementary information of the two 
kinds of data. The feature-level fusion methods\cite{AVOD-FPn,Huang2020EPNetEP,PIRCNN,Yoo20203DCVFGJ} generally adopt a two-branch structure for fusing image features and the point cloud features. However, it is difficult for existing methods to integrate multi-modal data well. First, due to the differences in the characteristics of different modalities, simply fusing them usually leads to unsatisfactory detection results. Second, directly fusing the two kinds of data involves processing the non-interested regions, which may have a negative impact on detecting objects.

\begin{figure}[t]
\centering
\includegraphics[width=\linewidth]{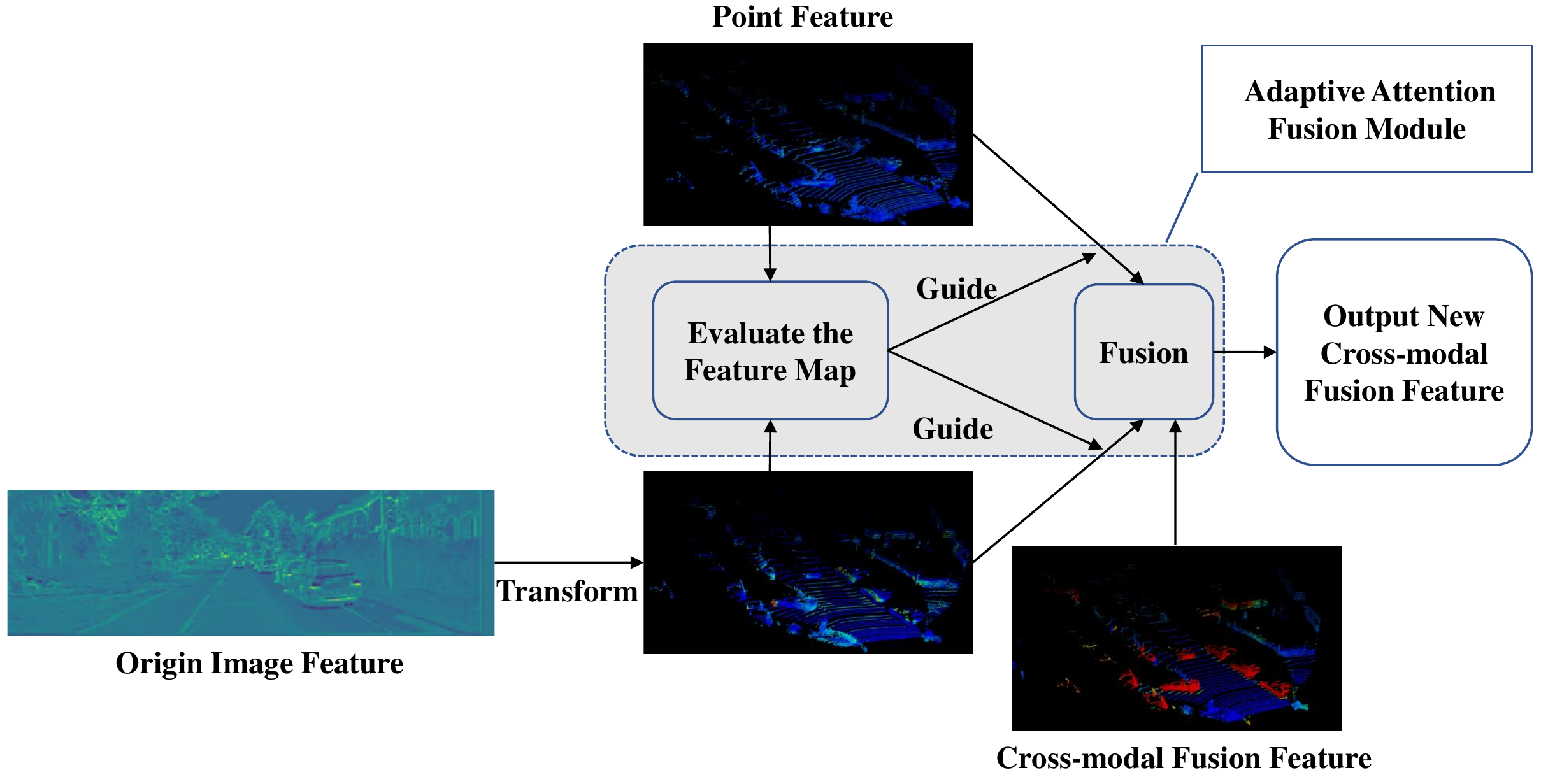}
\caption{Illustration of the fusion method for our cross-modal fusion feature. We present the workflow in AAF module.}
\label{intro}
\end{figure}

To solve this problem,  we propose a Multi-Branch Deep Fusion Network (MBDF-Net) for 3D object detection. Different from previous two-branch based methods, the proposed model consists of three branches, including an image branch, a LiDAR branch, and a fusion branch. The image branch and the LiDAR branch are used for extracting semantic information of the corresponding data modality. Specially, we utilize several Adaptive Attention Fusion (AAF) modules (see Figure \ref{intro}) to generate cross-modal fusion features for the fusion branch. AAF modules integrate the information of the two modalities and then evaluate each of them. In this way, the network can choose more essential information from two modality data and suppress the interference from non-interest areas. Once the 3D proposals are generated based on the cross-modal fusion feature, we utilize a region of interest (RoI)-pooled fusion module to produce enhanced local features for them. This module will enlarge the proposals and aggregate additional information, including the raw point coordinates, specific single-modal features, and cross-modal fusion features. The output local features can lead to more accurate refining results.

\begin{figure*}
\centering
\includegraphics[width=\linewidth]{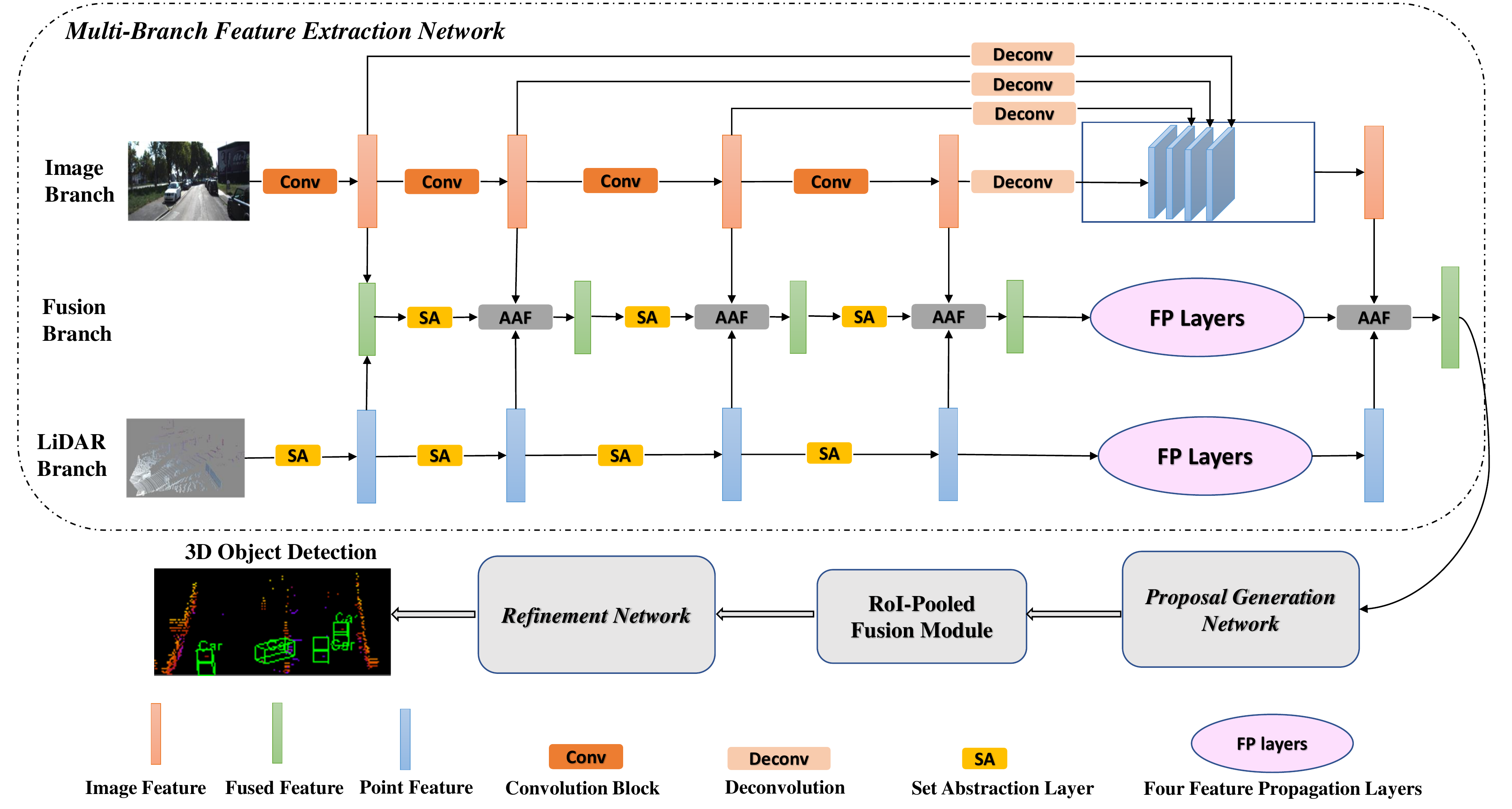}
\caption{Overview of the proposed MBDF-net structure. First, we extract semantic information from each modality and fuse them to generate cross-modal fusion features by AAF modules. After that, we use the RoI-pooled fusion module to produce enhanced local features for proposal refinement. In SA layers we replace FPS with hybrid sampling.}
\label{f2}
\end{figure*}

In addition, we notice that most current methods \cite{Qi2017PointNetDH, Shi2019PointRCNN3O, Huang2020EPNetEP} need to downsample points and then use the sampled points as the centers for extracting range features. In these methods, farthest point sampling (FPS) \cite{Qi2017PointNetDH} is often used. It considers the spatial information to ensure the uniformity of sampling. In this case, all regions have the same importance. However, in the object detection task, we believe it is necessary to pay more attention to the regional features of the objects. Moreover, if the selected point is close to the center of the object, extracting the range information of this point will help us to get more accurate location results. In order to select points that balance uniformity and representation in the downsampling procedures, we proposed a novel hybrid sampling method. First, we use the FPS to sample more points than the required number. Then, we utilize the point cloud attention map in the AAF modules to guide us to choose more critical points from them.
The proposed sampling method not only considers the information of the point itself but also integrates the information of the point and image.

Our contributions can be summarized as follows:
\begin{itemize}
\item We propose a MBDF-Net for 3D object detection. In the first stage, we utilize AAF modules to produce cross-modal fusion features. In the second stage, we generate enhanced local features by an RoI-pooled fusion module for better proposal refinement.
\item A novel hybrid sampling method is applied in point downsampling, which can not only guarantee the uniformity of sampling points but also choose more representative points.
\item We evaluate our methods on both the outdoor KITTI \cite{Geiger2013VisionMR} dataset and the indoor SUNRGBD \cite{sunrgbd} dataset. We achieve state-of-the-art performance.
\end{itemize}

\section{Related work}
\subsection{3D object detection based on LiDAR}

Currently, the LiDAR-based detection algorithms can be roughly grouped into two lines. The first kind of methods \cite{2017Mv3d,Yan2018SECONDSE, Kuang2020VoxelFPNMV, Zhu2019ClassbalancedGA, parta2, Ku2018Joint3P, Yang2018PIXORR3} usually use a convolutional neural network to extract features of projected of point clouds. For example, MV3D \cite{2017Mv3d} projects point clouds to bird’s eye view (BEV) perspective and generates bounding boxes with 2D convolution. MV3D \cite{2017Mv3d}, SECOND \cite{Yan2018SECONDSE}, voxel-net \cite{Kuang2020VoxelFPNMV} builds the voxels of the point clouds and then processes it with 3D convolution. Although convolution has better perception ability. It needs the pre-processing procedure and takes more time during calculation. Another kind of methods such as PointRCNN\cite{Shi2019PointRCNN3O}, STD\cite{Yang2019STDS3}, 3D-SSD\cite{Yang20203DSSDP3}, voxel-FPN\cite{Kuang2020VoxelFPNMV} directly deal with the original point clouds. PointRCNN \cite{Shi2019PointRCNN3O} directly uses PointNet++ \cite{Qi2017PointNetDH} as the feature extraction network to generate 3D recommendation boxes by dividing the foreground points of the point clouds. Recently, the two methods are combined in PV-RCNN \cite{Shi2020PVRCNNPF}. The point-based method is used to process the whole, and convolution is used to further extract the features in the proposal.

\subsection{3D object detection based on multiple sensors}
There are also a number of methods focusing on fusing the information of different sensors for 3D object detection \cite{Qi2018FrustumPF, pointfuse, Liang2020RangeRCNNTF, 2017Mv3d,AVOD-FPn, Contfuse,Huang2020EPNetEP, Yoo20203DCVFGJ, Pang2020CLOCsCO}. Previous methods such as Frustum PointNet \cite{Qi2018FrustumPF} first detect the targets in the image and then detect the targets in the corresponding point clouds region. In these methods, the detection performance heavily depends on the accuracy of 2D detection results. CLOCs \cite{Pang2020CLOCsCO} is proposed to use pre-trained 2D and 3D detectors directly in the form of late-fusion, making the proposals in different modalities connected, but not integrating the features. Some methods such as MV3D \cite{2017Mv3d}, AVOD \cite{AVOD-FPn}, and CONTFUSE \cite{Contfuse} transform the disordered point cloud into a structured point cloud, then fuse the features of multiple perspectives with the images. However, when building multiple perspectives, many resize and crop operations are involved, which has a risk of losing the spatial information. EPNET \cite{Huang2020EPNetEP} utilizes images to guide the feature fusion, which does not require additional annotation information. 3D-CVF \cite{Yoo20203DCVFGJ} aggregates the features of multi-view images and uses an automatic calibration module to reduce errors caused by projection. Currently, most of the feature fusion methods use a structure of a two-branch network. Compared with our multi-branch structure, we create intermediate features to integrate useful information. %Besides, we notice the sampling process during the point feature extraction and guide it with multi-modal information. 

\section{Method}
The proposed MBDF-Net aims to improve the effect of multi-modal fusion. It is a two-stage object detector. In stage-1, the network extracts key information from each data modality and fuses them through AAF modules for generating 3D RoI proposals. Then in stage-2, we adopt an RoI-pooled fusion module to generate enhanced local features for further refining the generated proposals.  The whole framework of MBDF-Net is presented in Figure \ref{f2}.

\subsection{Multi-Branch Feature Extraction Network}
Our feature extraction network consists of three branches, including an Image branch, a LiDAR branch, and a fusion branch.

\begin{figure}[t]
\centering
\includegraphics[width=\linewidth]{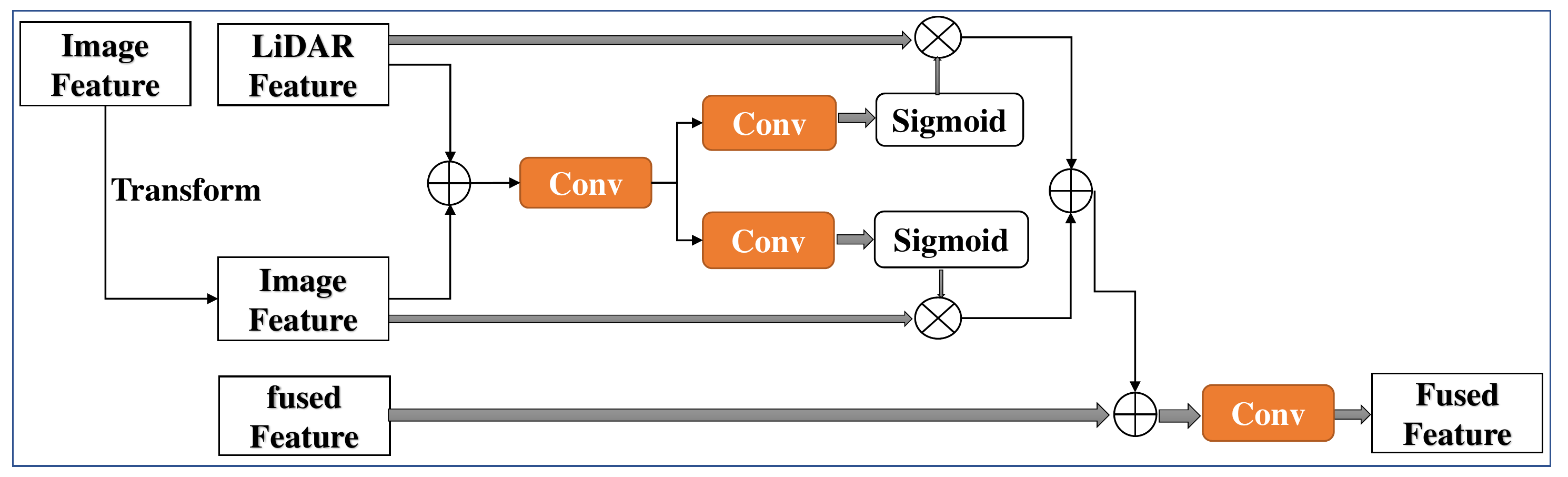}
\caption{Illustration of the AAF module. We first use projection to build the point-to-pixel relationship between image and point. So the transformed image feature can concatenate with the point feature in the same space. Then we concatenate them and feed them to the respective convolution layer followed by a sigmoid function. Finally, fused feature is generated by encoding all branch features. $\bigoplus$ denotes concatenation operation and $\otimes$ denotes multiply operation.}
\label{f3}
\end{figure}

{ \bf Image Branch} The image branch is used for extracting features from  images. It consists of four convolution blocks. Each convolution block is composed of a convolution layer with kernel size $3\times3$, a Batch Normalization layer, and a ReLU activation layer. The size of intermediate features produced by each block is reduced by two times to expand the receptive field. To use the detailed spatial information in low-level features, we use deconvolution to restore the feature of each block to the original image size and then concatenate them. We denote the final image feature as $F_{image}$.

{ \bf LiDAR Branch} The LiDAR branch is used for extracting features for point clouds. It utilizes the PointNet++ \cite{Qi2017PointNetDH} with multi-scale grouping as the backbone network to learn sophisticated and compact point-wise features. It contains four Set Abstraction (SA) layers and four Feature Propagation (FP) layers. Each SA layer first downsamples the point clouds and then extracts the local features. The FP layers are used to upsample the features to recover the original size of the point cloud. We denote the final point feature as $F_{point}$.

{\bf Fusion Branch} The fusion branch is used to extracting cross-modal fusion features and produce the initial 3D bounding box. Its input is generated by integrating the output features of the first convolution block and first SA layer from the image branch and the LiDAR branch respectively. We remove the first SA layer of PointNet++ \cite{Qi2017PointNetDH} and use the rest as its backbone network. As shown in Figure \ref{f2}, before feeding to each SA layer, we use AAF module to integrate the key information of the point and image features with the current cross-modal fusion feature. Through this method, the generated fused features contain rich point-wise semantic information. In the end, we utilize AAF module again to fuse $F_{image}$, $F_{point}$, and the output of this branch to get the final fused feature $F_{fused}$ for 3D proposal generation.

{\bf Adaptive Attention Fusion Module} In order to capture the key information of each data modality, we design an Adaptive Attention Fusion (AAF) module. The structure is shown in Figure \ref{f3}.

To fuse the two different views of data, we first use a projection method to establish the relationship between the points and the image pixels. Specifically, we multiply each point coordinate $p (x, y, z)$ with a projection matrix $M$ to generate the corresponding image coordinate $p' (x', y')$. $M$ is a inner parameters of the camera. Then we calculate the corresponding features of the point on the image by using bi-linear interpolation, as the projected points could fall between adjacent pixels. Finally, we produce the transformed image feature by assigning the features. 

It is difficult to fuse these two data because both are disturbed by non-interest areas. So we utilize attention mechanism to suppress the interference and extract essential information for fusion. We use the following formulas to generate the new fused feature,

\begin{equation}
    F_{attention-I}=\sigma \left(Conv_{I}\left(F_{image}\oplus F_{point}\right)\right),
    \label{E1}
\end{equation}
\begin{equation}
    F_{attention-P}=\sigma \left(Conv_{P}\left(F_{image}\oplus F_{point}\right)\right),
    \label{E1}
\end{equation}
\begin{equation}
\begin{aligned}
    F_{fused}=Conv \big( \left( F_{image} \times F_{attention-I} \right) \\
     \oplus \left( F_{point} \times F_{attention-P} \right) \oplus F'_{fused} \big),
    \end{aligned}
    \label{E3}
\end{equation}
where $\bigoplus$ represents the concatenation operation, $\sigma $ represents the sigmoid function. $F_{attention-I}$ and $F_{attention-P}$ represent the attention maps that indicate the relative importance of images and points.
% Firstly, we utilize image feature  ($F_{image}$) to get the attention map of the image ($F_{attention-i}$), and point feature ($F_{point}$) will get the attention map of the point ($F_{attention-p}$). The attention map indicates the relative importance of the image and point. Then, we use $\times$ to calculate the critical information of each modal. Finally, we fuse them with previous step fused feature ($F'_{fused}$) to produce our new fused feature ($F_{fused}$).

\begin{figure}
\centering
\includegraphics[width=\linewidth]{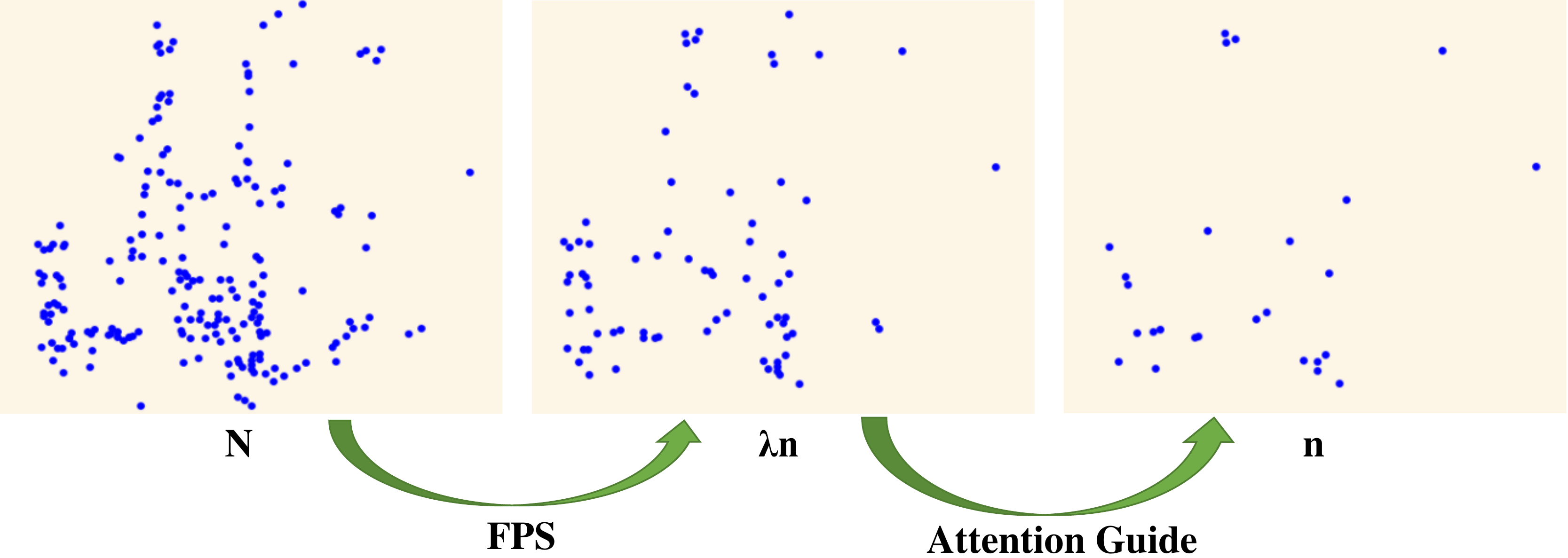}
\caption{Illustration of the hybrid sample strategy. We first use FPS to select $\lambda n$ from $N$ points. Then we choose $n$ points by utilizing attention to guide.}
\label{f10}
\end{figure}

\subsection{Hybrid Sampling Strategy}
%Of course, we experiment on the $\lambda$ value in the subsequent experiments and finally choose $\lambda$ = 1.4.

In the SA layers, we need to downsample points to expand the receptive field. The point cloud is disordered in space. To ensure the uniformity of sampling, the FPS is often used for covering the whole space as much as possible. However, The FPS only considers the relative distance between points without utilizing the rich information in the semantic features.
Using attribute features is another way to downsample points. It also has some problems. First, the selected points usually concentrate on the region of interest, which results in a serious aggregation phenomenon. Second, paying too much attention to the foreground point will decrease the ability to distinguish background points. Therefore, we present a hybrid sampling strategy (see Figure \ref{f10}) which takes both distance and attributes into consideration. 

Supposing that we need to sample $n$ points from $N$ points. First of all, we use the FPS to select $\lambda n (1.0 \leq \lambda \leq \frac{N}{n})$ points to ensure the uniformity of our sampling distribution. Then we get the attention map of the point $F_{attention-P}$ from AAF module. This map is produced under the guidance of point cloud and image features and represents important semantic information for each point. We get these $\lambda n$ attention scores from the $F_{attention-P}$. Next, these scores are sorted from large to small, and the $n$ points with higher scores are selected as the downsampling points. In this way, we select the key points with better attributes from the evenly distributed points and takes multi-modal information into account.

\subsection{RoI-Pooled Fusion Refinement Network}

\begin{figure}
\centering
\includegraphics[width=\linewidth]{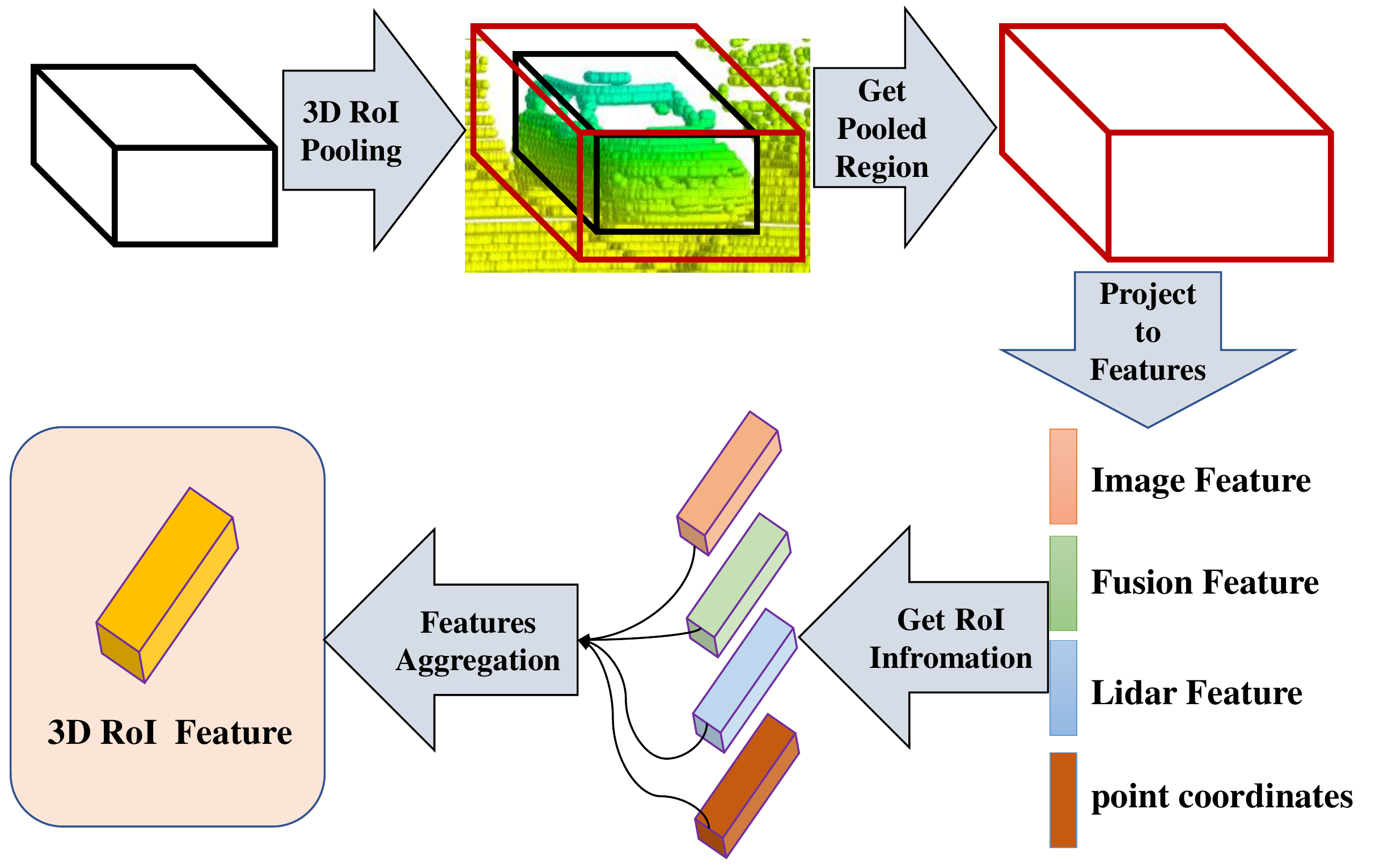}
\caption{Illustration of the RoI-pooled fusion structure. First, we enlarge the size of the RoI. Then we concatenate additional information in this area to produce the enhanced local feature. So abundant information is provided to the follow-up network. The green box is the original range size, and the red box is the expanded range size. }
\label{f5}
\end{figure}

In the first stage, the proposal generation network gets the initial regression results and corresponding scores. We use the non-maximum suppression (NMS) procedure to retain the proposals with high scores. In the second stage, to produce more precise results, firstly we use the RoI-pooled fusion module (see Figure \ref{f5}) to enhance each proposal.

We Keep the coordinates unchanged and expand the width, the length, and the height of the generated proposal, that is, from $(W, l, H)$ to $(W + a, L + a, H + a)$. $a$ is the parameter for enlarging the size of the proposals. Expanding the proposals could provide additional information for increasing the robustness. Then we aggregate the raw point coordinates, the specific single-modal features, cross-modal fusion features in each enlarged proposal. In this way, we can encode local spatial and semantic information to produce more rich local features for proposals. Finally, we feed these features to the refinement Network for refining the 3D bounding box.

\subsection{ Training Objective}
Our MBDF-Net is trained via a two-stage process. We use RPN loss to train the first stage and RCNN loss for the second stage. Each stage adopts a similar optimizing goal, includes a classification loss, and a regression loss. The overall objective is defined as:

\begin{equation}
\begin{aligned}
    L_{total}=&\left(L_{RPN-cls}+L_{RPN-reg}\right)+ \\
    &\left(L_{RCNN-cls}+L_{RCNN-reg}\right)
    \end{aligned}
    \label{E4}
\end{equation}
where $L_{RPN-cls}$, $L_{RPN-reg}$ indicate the RPN classification loss and regression loss respectively. The RCNN loss is similar.

To solve the problem that the proportion of positive and negative samples is too large, we use focal loss \cite{2017Focal} as the loss of classification. $L_{cls}$ is given by Eq.(5), 

\begin{equation}
    L_{cls}=-\alpha \left(1-c_{t}\right)^{\gamma }\log c_t
    \label{E5}
\end{equation}
where $c_t$ means the probability of the point belongs to the ground truth category, and we set $\alpha$ = 0.25 and $\gamma$ = 2.0. 

\begin{table*}
\caption{Performance comparison on the KITTI test set. The 3D and BEV results are evaluated by the mean Average Precision with 40 recall positions.}
    \begin{tabular}{ccccccccccccl}
    \toprule
    \multicolumn{1}{c}{\multirow{2}[0]{*}{Method}} & \multicolumn{1}{c}{\multirow{2}[0]{*}{Reference}} & \multicolumn{1}{c}{\multirow{2}[0]{*}{Modality}} & \multicolumn{4}{c}{3D Detection} & \multicolumn{4}{c}{Bird's Eye View } \\
          &       &       & \multicolumn{1}{c}{Easy} & \multicolumn{1}{c}{Moderate} & \multicolumn{1}{c}{Hard} & \multicolumn{1}{c}{3D mAP} & \multicolumn{1}{c}{Easy} & \multicolumn{1}{c}{Moderate} & \multicolumn{1}{c}{Hard} & \multicolumn{1}{c}{BEV mAP}\\
    \midrule
    \multicolumn{9}{l}{\textbf{LiDAR-only:}} \\
    % SECOND\cite{Yan2018SECONDSE} & \multicolumn{1}{l}{Sensors 2018} & Voxel & 83.34  & 72.55  & 65.82 & 73.90 & 89.39  & 83.77  & 78.59 & 83.92\\
    PointPillars\cite{2019PointPillars} & \multicolumn{1}{l}{CVPR 2019} & Voxel & 82.58  & 74.31  & 68.99 & 75.29 & 90.07  & 86.56  & 82.81 & 86.48\\
    % STD\cite{Yang2019STDS3}   & \multicolumn{1}{l}{ICCV 2019} & Voxel+Point & 87.95  & 79.71  & 75.09 & 80.91  & 94.74  & 89.19  & 86.42 & 90.12  \\
    PointRCNN\cite{Shi2019PointRCNN3O} & \multicolumn{1}{l}{CVPR 2019} & Point & 86.96  & 75.64  & 70.70 & 77.77 & 92.13  & 87.39  & 82.72 & 87.41 \\
    Point-GNN\cite{2020Pointgnn} & \multicolumn{1}{l}{CVPR 2020} & Point & 88.33  & 79.47  & 72.29 & 80.03 & 93.11  & 89.17  & 83.90 & 88.73 \\
    3D-SSD\cite{Yang20203DSSDP3} & \multicolumn{1}{l}{CVPR 2020} & Point & 88.36  & 79.57  & 74.55 & 80.83  & 92.66  & 89.02  & 85.86 & 89.18  \\
    SA-SSD\cite{He2020StructureAS} & \multicolumn{1}{l}{CVPR 2020} & Voxel+Point & 88.75  & 79.79  & 74.16 & 80.90  & 95.03 & \textbf{91.03 } & \textbf{85.96} & \textbf{90.67}  \\
    \midrule
    \multicolumn{9}{l}{\textbf{Multi-Modality:}} \\
    % MV3D\cite{2017Mv3d}  & \multicolumn{1}{l}{CVPR 2017} & RGB+LiDAR & 74.97  & 63.63  & 54.00 & 64.20 & 86.62  & 78.93  & 69.80 & 78.45 \\
    % AVOD-FPN\cite{AVOD-FPn} & \multicolumn{1}{l}{IROS  2017} & RGB+LiDAR & 83.07  & 71.76  & 65.73 & 73.52 & 90.99  & 84.82  & 79.62 & 85.14 \\
    % ContFuse\cite{Contfuse} & \multicolumn{1}{l}{ECCV 2018} & RGB+LiDAR & 83.68  & 68.78  & 61.67 & 71.38 & 94.07  & 85.35  & 75.88 & 85.10 \\
    F-PointNet\cite{Qi2018FrustumPF} & \multicolumn{1}{l}{CVPR 2018} & RGB+LiDAR & 82.19  & 69.79  & 60.59 & 70.86 & 91.17  & 84.67  & 74.77 & 83.54 \\
    UberATG-MMF\cite{UberATG-MM} & \multicolumn{1}{l}{CVPR 2019} & RGB+LiDAR & 88.40  & 77.43  & 70.22 & 78.68 & 93.67  & 88.21  & 81.99 & 87.96 \\
    PointPainting\cite{2019PointPainting} & \multicolumn{1}{l}{CVPR 2020} & RGB+LiDAR & 82.11 & 71.70 & 67.08 & 73.63 & 92.45 & 88.11 &83.36 & 87.97\\
    PI-RCNN\cite{PIRCNN} & \multicolumn{1}{l}{AAAI 2020} & RGB+LiDAR & 84.37 &	74.82 &	70.03 & 76.41  & 91.44 & 85.81 & 81.00 & 86.08   \\
    EPNet\cite{Huang2020EPNetEP} & \multicolumn{1}{l}{ECCV 2020} & RGB+LiDAR & 89.91  & 79.28  & 74.59 & 81.23 & 94.22  & 88.47  & 83.69 & 88.79 \\
    3D-CVF\cite{Yoo20203DCVFGJ} & \multicolumn{1}{l}{ECCV 2020} & RGB+LiDAR & 89.20  & \textbf{80.05 } & 73.11 & 80.79  & 93.52  & 89.56 & 82.45 & 88.51  \\
    \midrule
    OURS  &       & RGB+LiDAR & \textbf{90.87 } & 80.00  & \textbf{75.04 } & \textbf{81.97 } & \textbf{95.36 } & 89.18  & 84.24 & 89.59\\
    \bottomrule
    \end{tabular}%
  \label{t1}%
\end{table*}%

For the regression, we need to regress center point, size, and orientation $(x, y, z, l, h, w, \theta)$. And for the relatively small range for the $y, h, w, l,$ we use a smooth L1 loss \cite{Ren2015FasterRT}. For the $x, z, \theta$, we adopt a bin-based regression loss \cite{Qi2018FrustumPF, Shi2019PointRCNN3O, Huang2020EPNetEP}. First, it splits several bins for each point area, then it will predict which bin the center point will fall in. Last it regresses the residual offset within the bin. Besides, we utilize IoU loss as a regularization term. The overall regression loss $L_{reg}$ can be formulated as

\begin{equation}
\begin{aligned}
    L_{reg}=&\sum_{u\in x,z,\theta }^{}E\left(b_u,\hat{b}_u\right)+\sum_{u\in x,y,z,h,w,l,\theta }^{}S\left(r_u,\hat{r}_u\right) \\
            &-ln\left(\frac{Area\left(P\cap G \right)}{Area\left(P\cup G \right)} \right)
    \end{aligned}
    \label{E6}
\end{equation}
where $E$ denotes the cross-entropy loss and $S$ means the smooth L1 loss. $\hat{b}_u$ and $\hat{r}_u$ indicate the ground truth bins and offsets respectively. $P$ and $G$ are the predicted bounding box and ground truth.

\section{Experiments}
% In our experiments, we evaluated the proposed approach on the widely used KITTI \cite{Geiger2013VisionMR} dataset and SUNRGBD \cite{sunrgbd} dataset.

% \begin{table}
% \caption{Comparison between the ideas in the proposed method on the KITTI val dataset. }
% \resizebox{linewidth}{!}{
% \begin{tabular}{ccccccl}
% \toprule
% Use Fusion & RoI-Pooled Fusion & Hybrid Sample & Easy  & Moderate & Hard  & mAP   \\
% \midrule
% no & no & no & 91.20 & 79.47 & 77.27 & 82.65 \\
% yes & no & no & 92.01 & 82.23 & 79.90 & 84.71 \\
% yes & yes & no & 92.34 & 82.57 & 80.21 & 85.04 \\
% yes & yes & yes & 92.88 & 83.09 & 80.56 & 85.51 \\
% \bottomrule
% \end{tabular}
% }
% \label{t2}
% \end{table}

\subsection{Implementation Details	}
The proposed network takes both images and point cloud as input. The original resolution of images is $1280\times384$. We use four convolution blocks with stride 2 to downsample the image. Then, we employ four parallel transposed convolution with stride 2, 4, 8, 16 to recover the resolution from features in different scales, which is the same with EPNet \cite{Huang2020EPNetEP}. For each scene, we limit the input range of point cloud to (0, 70.4), (-40, 40), (-3, 1) in X, Y and Z directions respectively. We drop those points that are invisible in the image view. Then we subsample 16384 points from the raw point cloud as input. We follow the similar structure of PointNet++ \cite{Qi2017PointNetDH}, which contains four SA layers. We replace the sampling method (FPS) with our hybrid sampling method. We select $\lambda = 1.4$ and demonstrate the reasons in Section 4.4. The number of the point cloud sampled by the SA layers is $4096$, $1024$, $256$, and $64$ respectively. The original size of the point cloud is finally recovered by four FP layers. The settings of the fusion branch are the same as those of the point cloud branch. To remove the redundant proposals, we select the top 8000 boxes and set the non-maximum suppression (NMS) threshold to 0.8. Then we obtain 64 positive candidate boxes to refine. 

For the refinement network, we enlarge the size of each proposal by 0.2. Then, we sample 512 points from each proposal as input. For the proposals with less than 512 points, we supplement them with zeros. Next, we concatenate the raw point coordinates, each specific single-modal feature, and the cross-modal fusion feature for each proposal.
The generated features are fed into three SA layers with group size (128, 32, 1) to generate a feature vector for object confidence classification and location regression.

\begin{table*}
\caption{3D object detection results on SUN RGB-D V1 val set. We show the mean of average precision (mAP) across
10 classes with a 3D IoU threshold of 0.25. P and I represent the point cloud and the camera image.}
% \begin{center}
\begin{tabular}{ccccccccccccl}
\toprule
Method & Modality & bathtub & bed  & bookshelf & chair & desk & dresser & nightstand & sofa & table & toilet & 3D mAP   \\
\midrule
DSS\cite{dss} & P+I & 44.2 & 78.8 & 11.9 & 61.2 & 20.5 & 6.4 & 15.4 & 53.5 & 50.3 & 78.9 & 42.1 \\
2d-driven\cite{2ddriven} & P+I & 43.5 & 64.5 & 31.4 & 48.3 & 27.9 & 25.9 & 41.9 & 50.4 & 37.0 & 80.4 & 45.1 \\
COG\cite{cog} & P+I & 58.3 & 63.7 & 31.8 & 62.2 & \textbf{45.2} & 15.5 & 27.4 & 51.0 & 51.3 & 70.1 & 47.6 \\
PointFusion\cite{pointfuse} & P+I & 37.3 & 68.6 & \textbf{37.7} & 55.1 & 17.2 & 24.0 & 32.2 & 53.8 & 31.0 & 83.8 & 44.1 \\
F-PointNet\cite{Qi2018FrustumPF} & P+I & 43.3 & 81.1 & 33.3 & 64.2 & 24.7 & 32.0 & 58.1 & 61.1 & 51.1 & 90.9 & 54.0 \\
VoteNet\cite{votenet} & P & 74.4 & 83.0 & 28.8 & 75.3 & 22.0 & 29.8 & \textbf{62.2} & 64.0 & 47.3 & 90.1 & 57.7 \\
EPN\cite{DBC} & P+I & 79.4 & \textbf{88.2} & 32.1 & 17.0 & 37.4 & \textbf{53.7} & 50.0 & 65.3 & \textbf{53.3} & \textbf{95.8} & 57.2 \\
\midrule
Ours & P+I & \textbf{81.5} & 84.7 & 33.0 & \textbf{77.3} & 31.2 & 29.0 & 57.7 & \textbf{65.6} & 49.9 & 85.5 & \textbf{59.5} \\
\bottomrule
\end{tabular}
%\end{center}
\label{tsun}
\end{table*}

\begin{table*}
\caption{Comparison between the ideas in the proposed method on the KITTI val dataset. }
\begin{tabular}{ccccccl}
\toprule
Use Fusion & RoI-Pooled Fusion & Hybrid Sample & Easy  & Moderate & Hard  & mAP   \\
\midrule
no & no & no & 91.20 & 79.47 & 77.27 & 82.65 \\
yes & no & no & 92.01 & 82.23 & 79.90 & 84.71 \\
yes & yes & no & 92.34 & 82.57 & 80.21 & 85.04 \\
yes & yes & yes & 92.88 & 83.09 & 80.56 & 85.51 \\
\bottomrule
\end{tabular}
\label{t2}
\end{table*}

We train the network by the Adam \cite{2014Adam} optimizer. The initial learning rate, weight decay, and momentum factor are set to 0.002, 0.001, and 0.9 respectively. We use rotation around the vertical $Y$ axis between [-10, 10] degrees, random flipping, and scale transformations with a factor within [0.95, 1.05] on LiDAR point as data augmentation strategies to prevent over-fitting.

\begin{table}[t]
\caption{Comparison between the two branch network, three branch network, and three branch network with adaptive attention on KITTI val dataset.}
\begin{tabular}{ccccl}
\toprule
 Method & Easy  & Moderate & Hard  & mAP   \\
\midrule
Two Branch & 91.25 & 81.69 & 79.48 & 84.14 \\
Three Branch & 91.78 & 82.09 & 79.58 & 84.48 \\
Adaptive Attention & 92.01 & 82.23 & 79.90 & 84.71 \\
\bottomrule
\end{tabular}
\label{t3}
\end{table}

\subsection{KITTI}

\begin{figure*}[t]
\centering
\includegraphics[width=\linewidth]{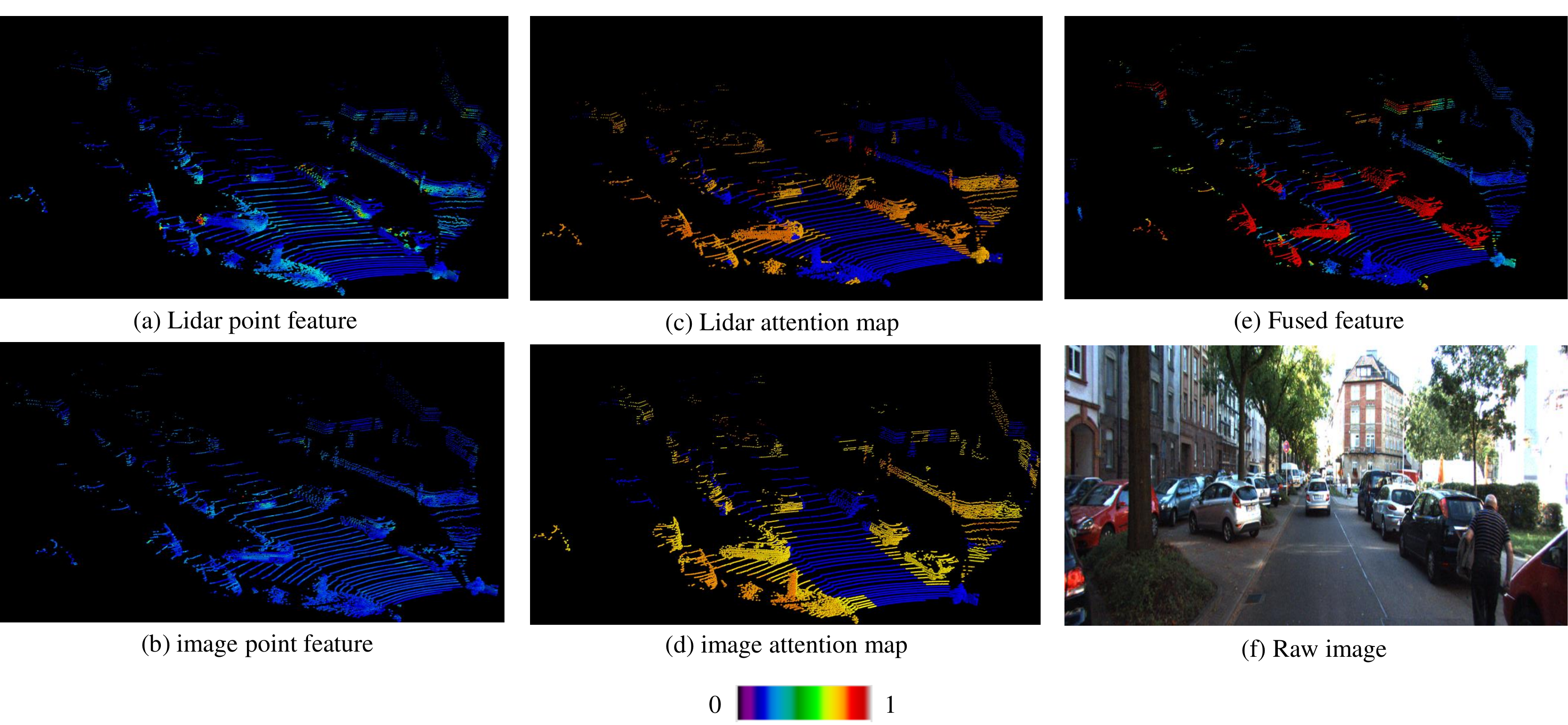}
\caption{Illustration of the learned features. (a) and (b) are the single-modal features. (c) and (d) are the learned attention map. (e) is the final produced fused feature. We show the raw image in (f).}
\label{f9}
\end{figure*}

{\bf Dataset} KITTI is a standard benchmark dataset for autonomous driving. It contains 7481 training samples and 7518 test samples. According to the size, occlusion, and truncation, objects are classified into different difficulty levels, including Easy, Moderate, and Hard. We follow the general split of 3712 training samples and 3769 validation samples. And we provide the results both on the validation set and the test set.

\begin{figure}[t]
\centering
\includegraphics[width=\linewidth]{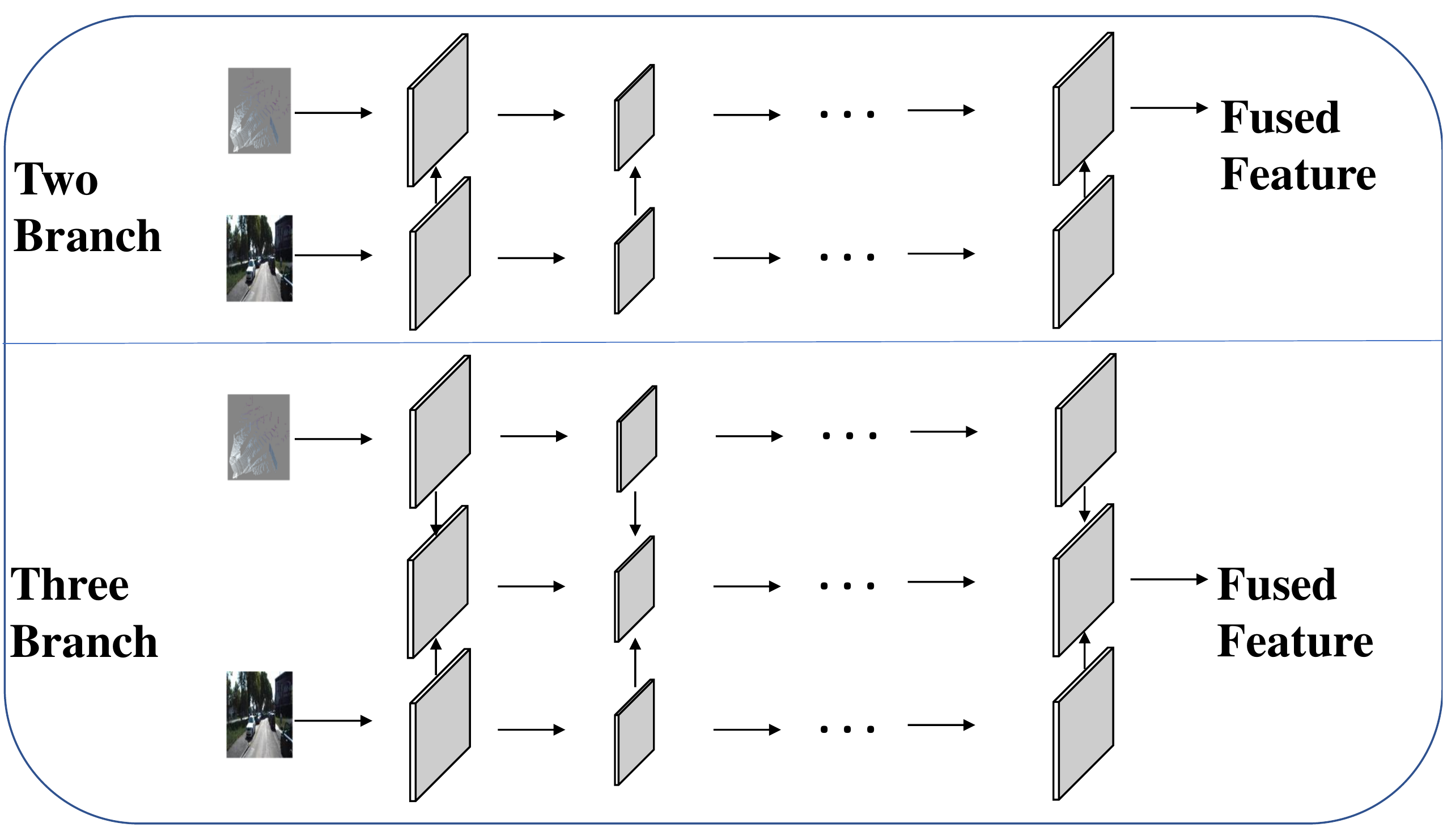}
\caption{Illustration of different designs of networks. Above is the widely used two-branch structure, Below is the three-branch structure we use. }
\label{f8}
\end{figure}

{\bf Metrics} The detection result is evaluated using the mean Average Precision (mAP) with the IoU threshold of 0.7. For the official test benchmark, the mAP with 40 recall positions is reported. For a fair comparison, we also report the mAP with 11 recall positions on the validation set.

{\bf Main Result} We present our results on KITTI benchmark in Table \ref{t1}. First, we compare our approach with a single-modal method PointRCNN \cite{Shi2019PointRCNN3O} since both methods use a similar backbone. Compared to \cite{Shi2019PointRCNN3O}, our approach gains huge improvements on 3D detection results by $3.91\%$, $4.36\%$, $4.34\%$, which correspond to the Easy, Moderate, and Hard difficulty respectively. Then, compared with multi-modal methods, our approach performs better than state-of-the-art methods MMF  \cite{UberATG-MM}, EPNet \cite{Huang2020EPNetEP}, and 3D-CVF  \cite{Yoo20203DCVFGJ} by $3.29\%$, $0.74\%$, $1.18\%$ on 3D mAP respectively. These methods use a two-branch structure. Specially, MMF\cite{UberATG-MM} improves 3D detection performance by using multiple auxiliary tasks such as 2D detection, ground estimation, and depth completion. 3D-CVF \cite{Yoo20203DCVFGJ} applies an auto-calibrated projection method to eliminate the error from projecting and utilizes a gated fusion network to aggregate the camera-LiDAR feature. EPNet \cite{Huang2020EPNetEP} integrates different features by adding the processed image information to the point feature. These results show the superiority of our methods both on single-modality and multi-modality approaches.

\subsection{SUN-RGBD}
{\bf Dataset} SUN RGB-D is a single-view RGB-D dataset for 3D scene understanding, which contains 10335 indoor RGB and depth images. We use the same training/validation split and report performance on the 10 most common categories as VoteNet \cite{votenet}. We use Average Precision(AP) and the mean of AP under IoU values of 0.25.

{\bf Main Result} The evaluation results on SUN-RGBD val set are shown in Table \ref{tsun}. Our MBDF-Net outperforms all previous methods. For 2D-driven \cite{2ddriven} and F-PointNet \cite{Qi2018FrustumPF} which is a 3D detector that based on 2D detection on images, we outperforms them by $14.4\%$ and $5.5\%$ mAP respectively. PointFusion \cite{pointfuse} combines image features and point features at the last high level. It lacks more accurate position information in low-level and does not consider the noise from the data. We outperform it by $15.4\%$. EPN \cite{DBC} is a two-branch detector. Our model improves by $2.3\%$ in mAP.
The results of indoor scenes also prove the effectiveness of our method.

\subsection{Ablation Study}
To give a deep insight into our approach, we conduct a number of experiments on the KITTI validation set.

\begin{figure}[t]
\centering
\includegraphics[width=\linewidth]{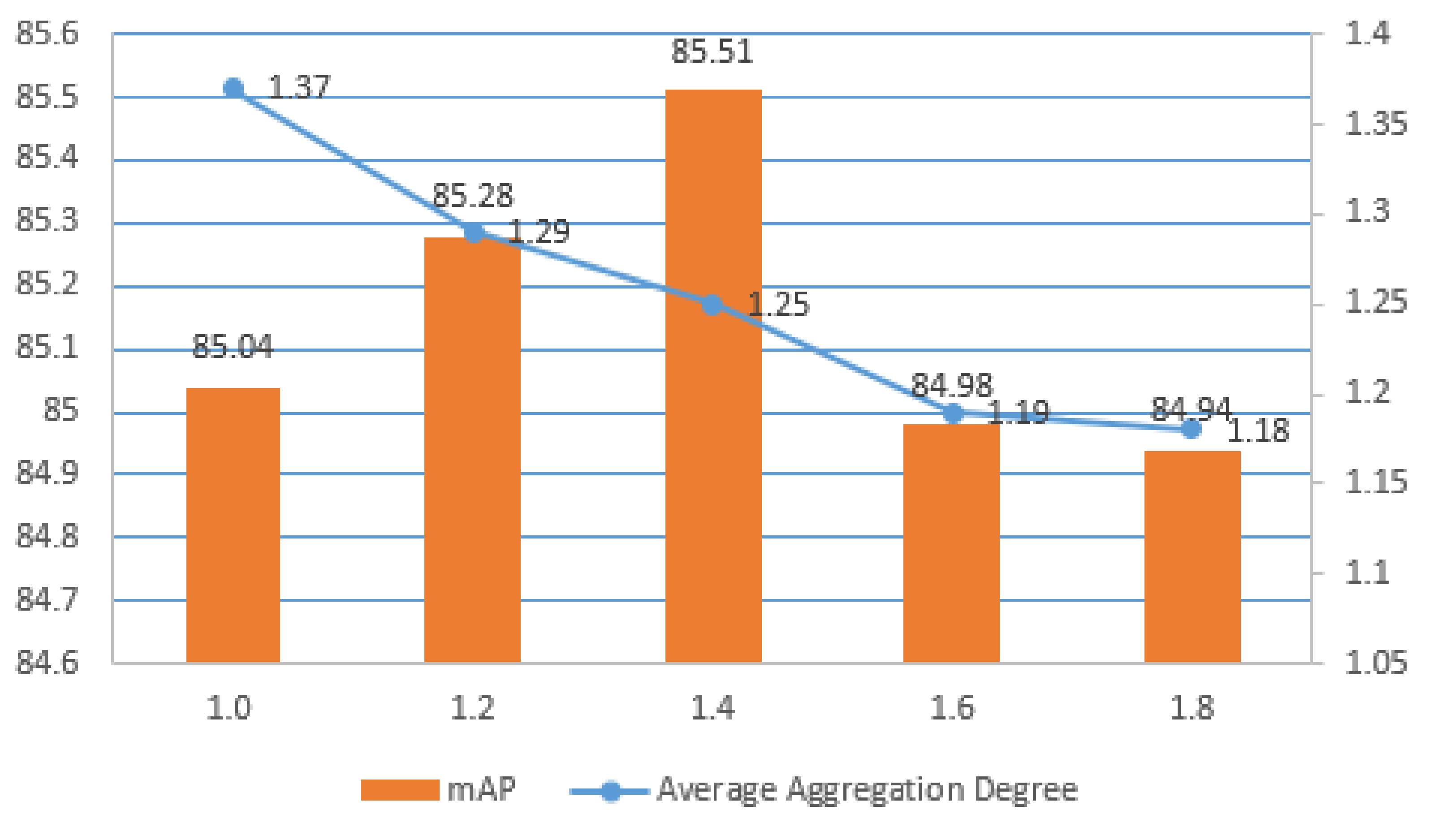}
\caption{Illustration of the average aggregation degree(AAD) and mAP with different $\lambda$ threshold.}
\label{f4}
\end{figure}

{\bf Analysis of Each Module:}
We present the result in table \ref{t2}. First of all, compared with the results of the pure point cloud, The accuracy of $AP_{easy}$, $AP_{moderate}$ and $AP_{hard}$ are improved by $0.81\%$, $2.76\%$ and $2.63\%$ respectively after using the multi-modal information. Next, when we use the RoI-pooled fusion module, our mAP is improved by $0.33\%$ due to more abundant information in the second stage. Finally, when we adopt the hybrid sampling method, we choose more excellent key points, which greatly improves our accuracy. The accuracy of $AP_{easy}$, $AP_{moderate}$ and $AP_{hard}$ are improved by $0.54\%$, $0.52\%$ and $0.45\%$ respectively. These results prove the effectiveness of proposed modules. 

\begin{figure}[t]
\centering
\includegraphics[width=\linewidth]{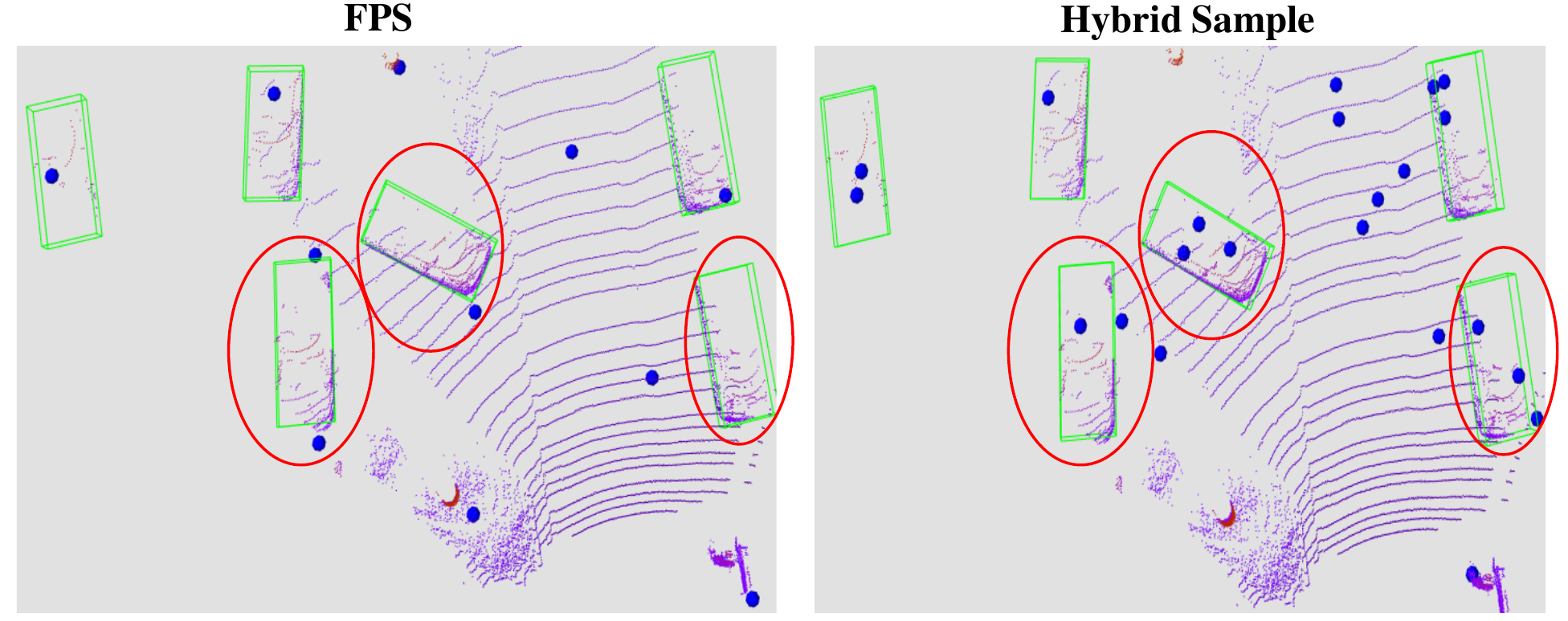}
\caption{Illustration of the distribution of different sampling methods. Ground-truth boxes are in green.}
\label{f11}
\end{figure}

{\bf Analysis of Fusion Strategies:}
We present the result in Table \ref{t3}. Firstly, to fairly compare with the widely used two-branch structure, we use simplified two-branch structure and our multi-branch structure (see figure \ref{f8}). In both structures, we just use concatenation to combine the different features. Our multi-branch structure is $0.34\%$ higher than the two-branch structure in $mAP$. Then, we utilize an adaptive attention mechanism to guide fusion, The accuracy was improved by $0.33\%$ $mAP$. From this results, we can see that the proposed multi-branch structure is better than widely used two-branch structure. Our fusion strategies can effective benefit the detection.

We further visualize the features in Fig \ref{f9}. (a) and (b) are the point features and image feature. (c) and (d) show that attention maps focus on the object areas and suppress the non-interested areas. (e) is the fused feature. 

{\bf Analysis of Hybrid Sampling Strategy:}

First, we define the average aggregation distance ($AAD$). For each point $P_{i}$, $AAD_{i}$ means the average distance from the three closest points to the $P_{i}$, which can be formularized as follows:

\begin{equation}
    AAD_{i}=\frac{1}{3}\sum_{j=1}^{3}\left(\left(x_{j}-x_{i}\right)^{2}+\left(y_{j}-y_{i}\right)^{2}+\left(z_{j}-z_{i}\right)^{2}\right)
    \label{E8}
\end{equation}

where $(x,y,z)$ is the central coordinates of the point. We calculate the average $AAD$ while utilizing our hybrid sampling strategy with different $\lambda$ in KITTI train set.  Specially, when $\lambda$ equals $1.0$, Our method is equivalent to FPS. We show the result in figure \ref{f4}. We can observe that with the increase of $\lambda$ value, the $AAD$ becomes smaller. This is easy to understand because points will cluster near the target. In other words, points will be unevenly distributed. In addition, the accuracy shows a trend of first rising and then falling. It rises from $85.04\%$ to $85.51\%$ and then falling to $84.95\%$. Combining the above phenomena, we can find that we need to choose the right $\lambda$ for optimal results. Because too small $\lambda$ only ensures that the points are evenly distributed. Too large $\lambda$ could lead points to concentrate in the region of interest and decrease the ability to distinguish background points. In the end, we choose $\lambda= 1.4$ , as it has a medium aggregation value of $1.25$, and optimal mAP $85.51\%$.

Table \ref{t6} shows the Comparison between the results of our network with farthest point sampling(FPS) and with attention-based hybrid sampling.
% To fully compare the effect between FPS and our hybrid sampling strategy. We test them in three benchmarks of the KITTI validation set(Cars). The results in Table \ref{t6} show that our attention-FPS improves the accuracy by $0.47\%$, $1.19\%$, and $0.09\%$ respectively. We also present the sample distribution in Figure \ref{f11}. When we use FPS, we can observe that there no points in some bounding boxes. After using our hybrid method, the sampled points can fall in the boxes and be close to the center.

\begin{table}[t]
  \caption{Comparison between the results of our network with farthest point sampling(FPS) and with attention-based hybrid sampling. We show the results on three benchmarks of the KITTI validation set(Cars). 3D, BEV, and AOS mean 3D Detection, Birds Eye View, and Orientation respectively.}
    \begin{tabular}{cccl}
    \toprule
    \multicolumn{1}{c}{\multirow{2}[0]{*}{Method}} & \multicolumn{3}{c}{mAP @0.5} \\
          & \multicolumn{1}{l}{3D} & \multicolumn{1}{l}{BEV} & \multicolumn{1}{l}{AOS} \\
    \midrule
    FPS   & 85.04 & 90.97 & 94.79 \\
    Attention-based Hybrid Sampling  & 85.51 & 92.16 & 94.88 \\
    \bottomrule
    \end{tabular}%
  \label{t6}%
\end{table}%

% \subsection{Qualitative Results}
% In Figure \ref{f6}, we present some qualitative Results for our proposed MBDF-Net on the KITTI test val.

\section{Conclusion}
In this paper, we present a new camera and LiDAR fusion two-stage framework MBDF-Net for 3D object detection. In the first stage, the point cloud branch, and image branch are utilized to obtain high-quality single-modal feature information. We use the AAF module to capture the key information of single-modal and produce rich semantic fused features. In the second stage, we use RoI-pooled fusion module to expand the range of the bounding box and generate enhanced local feature by providing additional information. In addition, we propose a novel hybrid sampling method, which can not only ensure the uniformity of spatial distribution of sampled points but also evaluate the values of the points themselves for subsequent processing. 
% The experimental results show that our model can achieve excellent performance among existing algorithms. 

\section{Acknowledgements}
This work was supported in part by NSFC under grant  No.621 2500145, No.62088102, No.61973246, No.91748208, Shaanxi Project under grant No.2018ZDCXLGY0607, and the program of the Ministry of Education.

\bibliographystyle{ACM-Reference-Format}
\bibliography{cite}
\end{document}